\documentclass[a4paper,conference]{IEEEtran}

\usepackage[utf8]{inputenc}
\usepackage{hyperref}
\usepackage{amsmath}
\usepackage{graphicx}
\usepackage{tabularx}
\usepackage[flushleft]{threeparttable}

\DeclareMathOperator*{\argmin}{arg\,min}

\ifCLASSINFOpdf
\else
\fi

\begin{document}
%
\title{High-resolution Ecosystem Mapping in Repetitive Environments Using Dual Camera SLAM}

\author{\IEEEauthorblockN{Brian M. Hopkinson}
\IEEEauthorblockA{Department of Marine Sciences,
University of Georgia\\
Athens, Georgia 30602, USA\\
Email: bmhopkin@uga.edu}
\and
\IEEEauthorblockN{Suchendra M. Bhandarkar}
\IEEEauthorblockA{Department of Computer Science,
University of Georgia\\
Athens, Georgia 30602, USA\\
Email: suchi@uga.edu}
}

\maketitle


\begin{abstract}
Structure from Motion (SfM) techniques are being increasingly used to create 3D maps from images in many domains including environmental monitoring. However, SfM techniques are often confounded in visually repetitive environments as they rely primarily on globally distinct image features.  Simultaneous Localization and Mapping (SLAM) techniques offer a potential solution in visually repetitive environments since they use local feature matching, but SLAM approaches work best with wide-angle cameras that are often unsuitable for documenting the environmental system of interest. We resolve this issue by proposing a dual-camera SLAM approach that uses a forward facing wide-angle camera for localization and a downward facing narrower angle, high-resolution camera for documentation. Video frames acquired by the forward facing camera video are processed using a standard SLAM approach providing a trajectory of the imaging system through the environment which is then used to guide the registration of the documentation camera images. Fragmentary maps, initially produced from the documentation camera images via monocular SLAM, are subsequently scaled and aligned with the localization camera trajectory and finally subjected to a global optimization procedure to produce a unified, refined map. An experimental comparison with several state-of-the-art SfM approaches shows the dual-camera SLAM approach to perform better in repetitive environmental systems based on select samples of ground control point markers. 
\end{abstract}

\begin{figure*}[htbp!]
\centerline{\includegraphics[width=\textwidth]{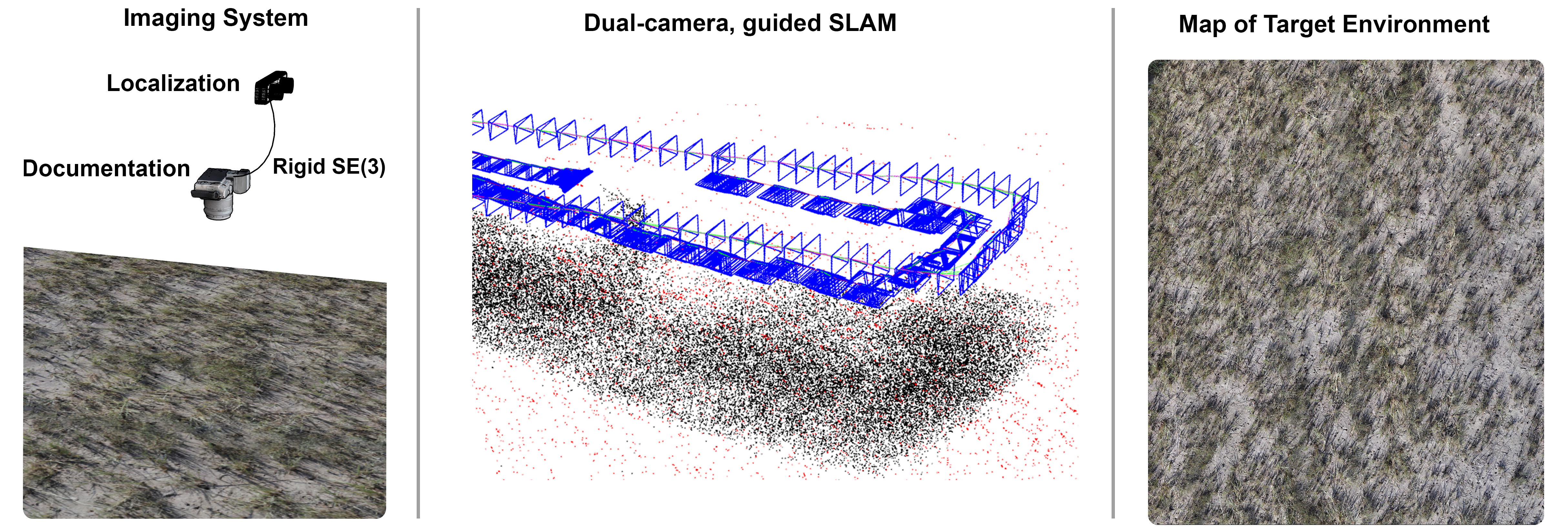}}
\caption{\textbf{Overview of Proposed Approach:} The imaging system consists of two cameras: a forward facing stereo-camera for localization and a downward facing, high-resolution camera for documentation. The dual-camera SLAM system produces trajectories (magenta line), landmark maps (black and red dots), and image poses (blue frames) for each camera. The image poses and landmark map from the documentation camera can be used to produce a map of the targeted, visually-repetitive environment as shown here for a salt marsh } 
\label{fig:Overview}
\end{figure*}

\vspace{-0.05in}
\section{Introduction}
\vspace{-0.05in}
The spatial arrangement of organisms within an ecosystem reflects fundamental underlying ecological processes such as competition, resource availability, trophic relationships, and symbioses~\cite{Levin_1992, Tarnita_2017}.  Consequently, the ability to map the abundance and distribution of organisms within an ecosystem is critical for advancing ecology. Traditionally, mapping organismal distributions entailed time-consuming, manual field work, limiting the scale and frequency at which maps could be generated. In recent decades, satellite remote sensing has offered unprecedented insight into the spatial arrangement and coverage of various ecosystems, but because of the coarse resolution of satellite imaging ($\sim$ 1-30 m pixel size) it is typically only useful at the ecosystem level (e.g. distribution of forest vs. grassland) and cannot assess the distribution of individual species~\cite{Schimel_2015}. Recent advances in computer vision algorithms, most notably in Structure-from-Motion (SfM) ~\cite{Fitzgibbon_1998, Snavely_2006, Agarwal_2009}, have begun to see their application in ecology for construction of much higher resolution (mm to cm) 3D optical maps of ecosystems. When combined with machine learning tools for automated classification, these maps are capable of delineating the distribution of individual species across landscape scales~\cite{Brodrick_2019, Hopkinson_2020}. 

SfM has been the tool of choice for map generation from images for ecologists and geographers due to its ability to use structured or unstructured image collections and the availability of high-quality commercial and open-source implementations~\cite{Schonberger_2016, Jackson_2020}. SfM relies on \textit{globally} distinct visual features in images to register overlapping images for map generation. This requirement is typically met when images are acquired at high elevations via unmanned aerial vehicles (UAVs or drones). However, when ecosystem images are acquired closer to the scene, for example to resolve small plants or animals, they often become repetitive causing conventional SfM approaches to fail. In contrast, Simultaneous Localization and Mapping (SLAM) approaches developed in the robotics community, process images sequentially in the order they were acquired using \textit{locally} distinct visual features to map the environment and localize the position and orientation of the camera with respect to the map~\cite{Durrant-Whyte_2006, Cadena_2016}.  SLAM approaches, though promising for mapping repetitive scenes ~\cite{Shu_2020}, work best with high-frame rate (and consequently lower resolution), wide-angle cameras whose images are of limited use for identification and localization of organisms~\cite{Davison2004}. 
	
In this paper, we describe a dual-camera SLAM approach to map visually repetitive environments such as grassland and shrubland ecosystems (Fig.~\ref{fig:Overview}). A high-frame rate, wide-angle camera is used for conventional visual SLAM-based localization whereas the other high-resolution, medium- to narrow-angle (video or still image) camera is used to acquire high-quality ecosystem images suitable for “documentation”, i.e., identification and localization of organisms to the species or genus level. The documentation camera does not need to be tightly integrated at the hardware level with the SLAM camera, allowing the use of extremely high-quality and low-cost commercial off-the-shelf (COTS) cameras. The trajectory of the localization camera is then used to guide detailed map generation from the documentation camera images using the proposed dual-camera SLAM approach. 

The primary contributions of this work are: (a) a novel approach to ecosystem map generation that allows flexible use of high-resolution, medium- to narrow-angle COTS cameras to resolve smaller organisms by decoupling localization and documentation; (b) development of a multistage alignment process for the documentation camera images that uses the localization camera to guide image pose determination and map point positioning; and (c) experimental demonstration of the proposed system's ability to map visually repetitive environments. 

\vspace{-0.05in}

\section{Related Work}

\vspace{-0.05in}

\subsection{Ecosystem Mapping}

Although mapping the distribution of ecosystems (e.g. forest, coral, and grassland extent) from satellite imagery has a long history~\cite{Schimel_2015, Anderson_2018}, we confine our overview to methods that exploit higher-resolution imagery (mm-cm pixel size) enabling more fine-grained taxonomic resolution (species, genera) since they are more closely related to our work. The most common ecosystem mapping workflow is comprised of image acquisition from a structured aerial survey using a UAV at $\approx$10-250 m elevation followed by image positioning and map generation via SfM. This approach provides ground resolution of $\approx$1 cm, suitable for classifying and delineating moderate- to large-sized organisms. The combination of UAVs, COTS cameras, and SfM has proved to be both, highly accurate and cost-effective avoiding the need for high-cost precise positioning sensors (e.g. RTK-GPS, IMUs) or high-cost imaging sensors (e.g. LiDAR, hyperspectral). 

Hayes et al.~\cite{Hayes_2021} use SfM to construct orthomosaics of seabird colonies from UAV images acquired at 60-90 m altitude followed by CNN-based object detectors to count individual birds for population tracking.  Baena et al.~\cite{Baena_2017} map the distribution of the keystone Algarrobo tree in Pacific Equatorial dry forests using SfM procedures on large-scale 260 m altitude UAV imagery. However, in low-visibility underwater environments, images are typically acquired closer to the scene (several meters) either manually or using underwater vehicles and then processed via SfM~\cite{Edwards_2017} or customized approaches~\cite{Bodenmann_2017}. 

\vspace{-0.05in}
\subsection{Structure from Motion (SfM)}

SfM approaches to scene reconstruction (i.e., mapping) and image pose determination are currently the primary tool for image-based ecosystem mapping. SfM attempts to map a scene from unordered images from uncalibrated and possibly multiple cameras, imposing minimal constraints on image acquisition~\cite{Agarwal_2009, Schonberger_2016}. The relative pose between images is computed by extracting features and associated descriptors (e.g. SIFT) from the unordered image collection and matching these features between images, using geometric verification to remove outliers~\cite{Hartley_MVG}. Matched feature points are triangulated producing a sparse representation of the scene. The image poses, scene points, and camera calibration parameters are optimized via a bundle adjustment procedure~\cite{Triggs_2000}. This sparse scene representation can further be made dense using multi-view stereo methods~\cite{Furukawa_2010}, and/or converted to a triangular mesh representation. 

 SfM is extremely flexible and relatively easy for non-specialists to use since it imposes minimal constraints on the image acquisition process, but the allowance for unordered image sets makes SfM computationally expensive with a non-linear time complexity ~\cite{Agarwal_2009, Wu_2013}. Furthermore, because images are unordered they must be visually similar only to their true neighbors; otherwise matches between spatially disparate locations may be incorrectly accepted.  Consequently, SfM is often confounded in scenes with repetitive features when this requirement is violated resulting in \textit{visual aliasing}~\cite{Lajoie_2019}.  

\vspace{-0.05in}
\subsection{Simultaneous Localization and Mapping (SLAM)}

SLAM techniques typically provide a similar final mapping solution to SfM, i.e.,  a sparse or dense representation of the scene and image poses~\cite{Cadena_2016}. However, SLAM assumes that the images are acquired sequentially, reflecting its origins in robotics. For ecosystem mapping, the sequential image acquisition constraint is generally not onerous as images are typically acquired from a single camera as it is moved over the underlying ecosystem. Since images are processed sequentially, the features need be only locally distinct, thus making SLAM better suited to handle repetitive environments. However, SLAM works best with a wide-angle, high frame rate camera whose limited image resolution is typically impractical for organismal identification~\cite{Davison2004}. This motivates our incorporation of an additional, higher resolution camera for ecosystem documentation. Although several multi-camera SLAM systems have been developed for robotics, they assume that precise synchronization information in the form of timing and orientation are available for each camera ~\cite{Urban_2017,Kuo_2020, Tribou_2015}, thereby preventing use of most inexpensive, high quality COTS cameras, which do not expose synchronization signals. By relaxing the precise synchronization requirement, we allow use of COTS cameras, but do not incorporate documentation camera images into the estimation of the system pose.

\vspace{-0.05in}
\section{Proposed System}
\begin{figure*}[htbp!]
\centerline{\includegraphics[width=\textwidth]{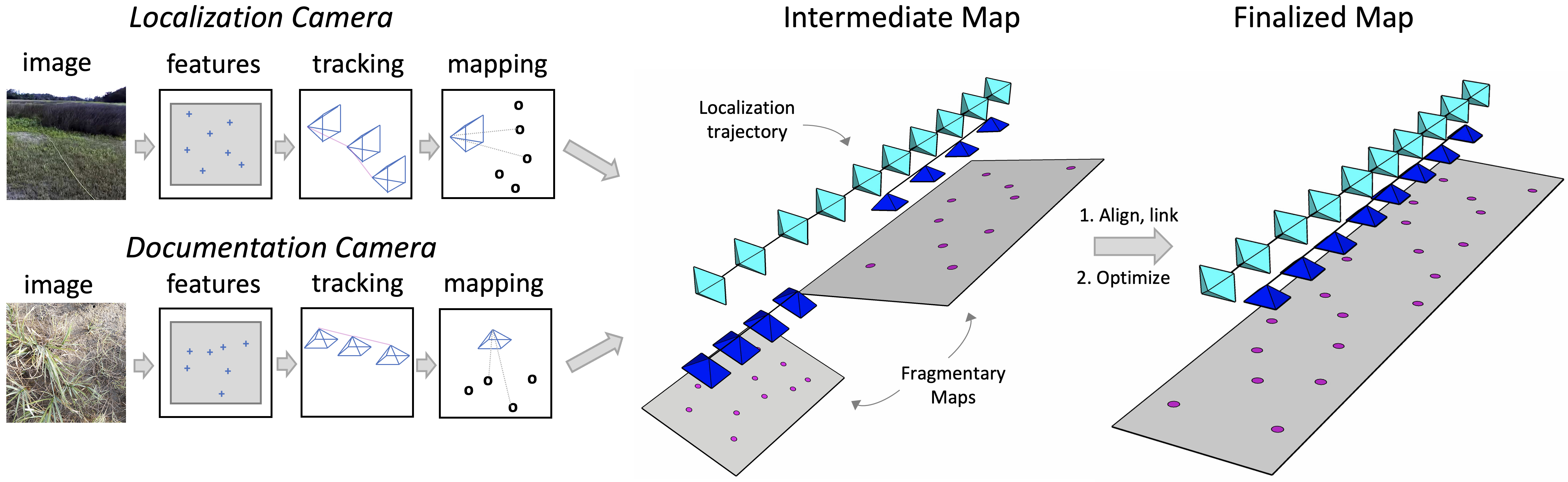}}
 \vspace{-0.1in}
 \caption{\textbf{Dual Camera SLAM Procedure:} Images from the localization and documentation cameras are processed in parallel starting with feature extraction, followed by initial pose estimation through landmark tracking, and finally optimization of the pose and generation of new landmarks in the mapping phase. Processing of the entire video streams results in an intermediate map consisting of a unified map for the localization camera and fragmentary maps from the documentation camera. The localization camera's trajectory is used to align the fragmentary maps, which are then linked through commonly viewed landmarks. Finally, all poses and landmarks are globally optimized to produce a finalized map from the documentation camera. } 
\label{fig:Architecture}
\end{figure*}

\vspace{-0.05in}
The proposed dual-camera SLAM-based ecosystem mapping approach uses SLAM to determine the trajectory of the localization camera, which is then used to guide map generation from the documentation camera images (Fig.~\ref{fig:Architecture}). The system assumes the relative orientation of the two cameras is constant and approximately known and that the image streams are roughly (\textless 0.5 s) synchronized. As SLAM is applied to the localization camera images, the documentation camera images are processed concurrently via monocular SLAM using the localization trajectory to guide generation of an initial ecosystem map and approximate image poses. However, the initial ecosystem map is generally fragmentary as tracking is frequently lost due to rapid scene movement resulting from the narrow field of view (FOV) and limited number of trackable features in the documentation camera images. After completion of the SLAM processing of the localization camera image sequence, the fragmentary maps from the documentation camera are scaled and transformed to approximately align with the localization camera trajectory based on the acquisition time of documentation camera images and the pose of the documentation camera relative to the localization camera. Finally, the documentation camera poses and associated map are optimized based on constraints derived from the localization camera trajectory and landmark-to-camera correspondences in the fragmentary maps using a factor graph framework.

\vspace{-0.05in}
\subsection{SLAM System Core}

The proposed system employs a modified version of ORB-SLAM2~\cite{Mur-Artal_2017}, a keyframe-based SLAM approach, as its base since it is a comprehensive SLAM approach capable of using monocular, stereo, and RGB-D cameras and produces accurate maps through multiple rounds of optimization (i.e., bundle adjustment). ORB-SLAM2 performs the following main operations: {\it tracking}, which localizes each frame relative to the existing map and determines when new keyframes should be added, {\it mapping}, which maintains the current map and updates it via insertion of new keyframes, triangulation of new map points, and optimization of the map structure via bundle adjustment, and {\it loop closing}, which identifies revisited locations (i.e., trajectory loops) and revises the map accordingly. 

We made several notable modifications to the ORB-SLAM2 system. First, the system was extended to handle multiple cameras simultaneously, with each camera maintaining a separate master map. The map data structure, which holds keyframes and map points, was converted into a recursive tree structure to handle sub-maps that are generated when tracking is lost. Sub-maps can be kept private or optionally registered with their parent to make keyframes and map points accessible to the parent. The \textit{relocalization} state, which ORB-SLAM2 enters immediately upon loss of tracking, was eliminated; instead a new sub-map is spawned and SLAM initialization started upon loss of tracking. For the localization camera, the new sub-map is registered with the parent assuming the camera maintained a constant velocity between the time when tracking was lost and when a new map was successfully initialized. This time interval is typically very short (\textless 0.1 s) but if initialization is not successful after a set number of frames the sub-map is not registered. The per-frame camera trajectories are explicitly recorded as $SE(3)$ transformations relative to the reference keyframes, whose positions are continuously updated via optimization. The availability these trajectories is critical for positioning of the documentation camera images based on the pose of the localization camera. Our SLAM code is available online at \url{ https://github.com/bmhopkinson/hyslam}.


\subsection{Generation of the Fragmentary Map}

The first step in generating the ecosystem map and associated documentation image poses is the application of the modified monocular ORB-SLAM2 system to the sequentially acquired documentation images. A new map is initialized by tracking ORB feature points through multiple frames until there is sufficient parallax to accurately triangulate map points corresponding to the tracked features. In our system, the resulting map of arbitrary scale is brought into a consistent scale (approximately) with the localization camera map by estimating the absolute motion of the documentation camera over the initialization period using the motion of the localization camera. The pose of the documentation camera can be estimated at any time $t$ as:
\begin{equation}
    \label{eq:DocPosfromSLAM}
    \textbf{X}_{d}(t) = \textbf{X}_{l}(t) \textbf{T}_{dl}
\end{equation}
\noindent 
where $\textbf{X}_{d}(t)$ is the pose of the documentation camera at time $t$, $\textbf{X}_{l}(t)$ is the pose of the localization camera at time $t$, and $\textbf{T}_{dl}$  is the rigid-body $SE(3)$ transformation between the localization and documentation cameras. All poses are expressed in the camera-to-world transformation convention. The motion of the documentation camera over the initialization period can then be determined as:
\begin{equation}
    \label{eq:VDocfromSLAM}
    \textbf{V}_{d0}(t) = \textbf{X}_{d0}^{-1}\textbf{X}_{d1}
\end{equation}
Equation~(\ref{eq:VDocfromSLAM}) is necessary because although both cameras are rigidly attached to a frame, the motion experienced by each camera may be different. For example, pure rotation about an axis of the localization camera will induce a rotation and translation of the documentation camera when the documentation camera is spatially offset relative to the axis of rotation. The new map is scaled using the estimated motion ($\textbf{V}_{d0}$), which works well in most cases but can occasionally be inaccurate as a result of small absolute distances traveled during monocular SLAM initialization. The fragmentary maps are later rescaled using the full distance travelled during their generation providing a consistent, absolute scale among the fragmentary maps. 

Once the map is initialized, the documentation images are processed using standard monocular SLAM~\cite{Mur-Artal_2015,Mur-Artal_2017} which provides a convenient and robust way to determine approximate relative poses between documentation images and identify features that are consistently matched and geometrically verified. Tracking is often lost due to the narrow FOV and rapid relative motion of the downward facing documentation camera at which point a new fragmentary map is initialized. 

\vspace{-0.05in}
\subsection{Alignment of Fragmentary Maps}

The relevant outputs of the initial processing steps consist of the full localization camera trajectory and multiple fragmentary maps from the documentation camera as depicted in the Intermediate Map in Fig.~\ref{fig:Architecture}. The fragmentary maps are approximately scaled to the localization trajectory but their orientations and positions may diverge substantially from their true values. Although a full non-linear optimization procedure is ultimately used to provide the best estimate for the documentation camera map, proper initialization is necessary for convergence. Since the documentation camera sub-maps often deviate substantially from their correct configurations, a two-step procedure is used to approximately align the fragmentary maps with the localization camera trajectory. First, the camera centers for the documentation images in each fragmentary map are brought into alignment with their expected positions based on the localization camera trajectory using a $Sim(3)$ transformation estimated using Horn’s method~\cite{Horn_1987}. Since the path traveled within any fragmentary map is often approximately linear, there is a rotational ambiguity in the documentation camera poses when aligned based solely on the camera centers. To align the documentation image orientations an optimal $SO(3)$ transform is determined, again using Horn’s method~\cite{Horn_1987}, from the camera poses augmented with points representing unit positions along the pose axes. The $SO(3)$ transform is applied to the documentation camera poses, resulting in fragmentary maps in a coherent world coordinate system defined by the localization camera. The two-step procedure ensures that the arbitrary-length vectors taken to represent positions along the camera frame axes used to resolve the rotation ambiguity do not influence the scale estimated in the first step.  

\vspace{-0.05in}
\subsection{Linking Fragmentary Maps}

To provide inter-map constraints, landmarks viewed in multiple fragmentary maps are identified and redundant landmarks removed. For each fragmentary map, landmarks from other sub-maps potentially visible in the fragmentary map’s keyframes are collected. Correspondences between these landmarks and keypoints in the keyframes are determined and validated using geometric and feature-based criteria. When sufficient correspondences are established to keypoints associated with a landmark in the current fragmentary map the corresponding landmarks are merged.  

\subsection{Global Optimization of the Ecosystem Map}

\begin{figure}[]
  \includegraphics[width=\linewidth]{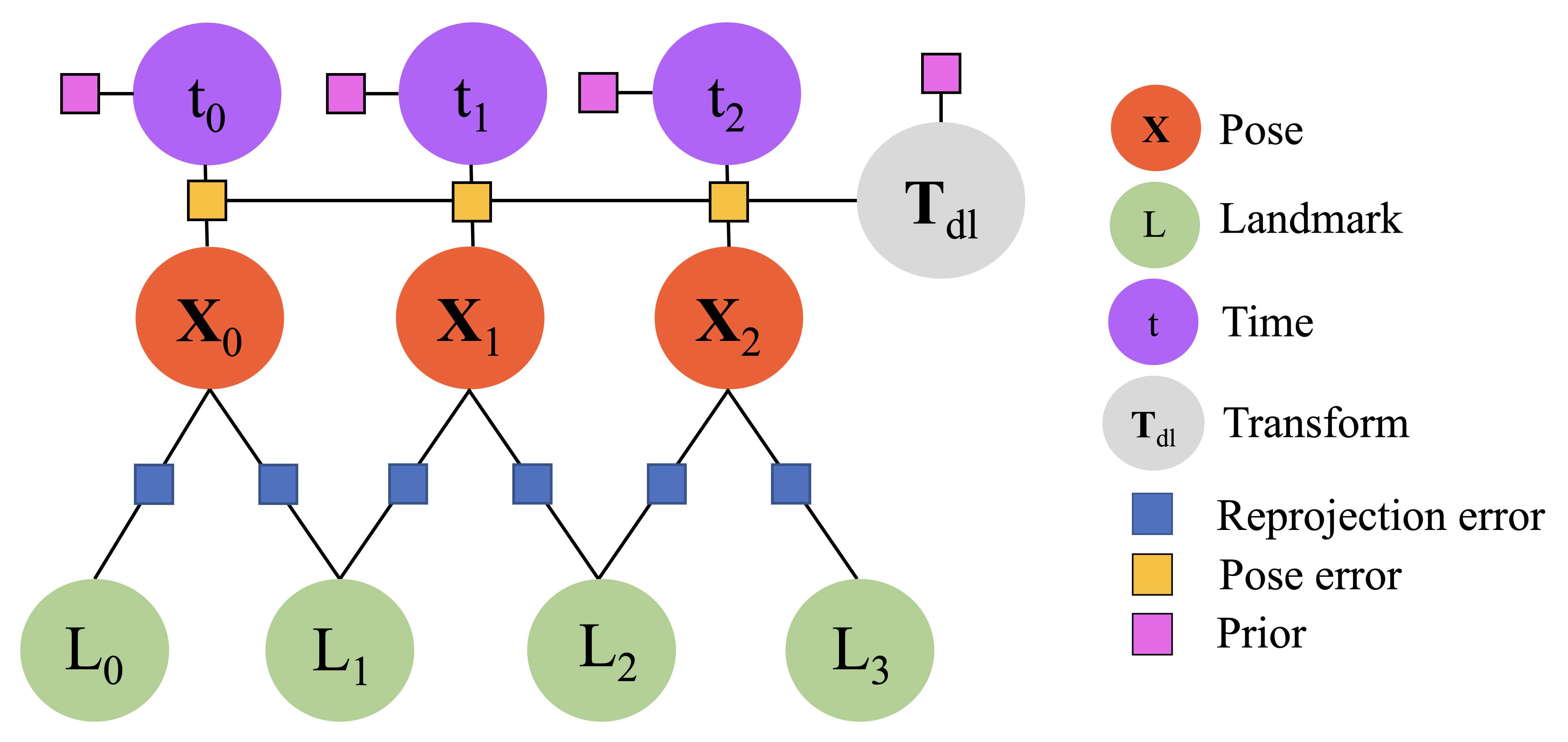}
  \vspace{-0.25in}
  \caption{\textbf{Documentation Camera Factor Graph}:  Circles represent model parameters estimated via global optimization and squares represent error terms constraining the parameters.} 
  \label{fig:Factor_Graph}
\end{figure}

The previously described procedures yield a set of fragmentary ecosystem maps from the documentation camera images approximately aligned based on the localization camera trajectory and linked via mutually visible landmarks. This ecosystem map, comprising of keyframes and landmarks, is refined using global, non-linear optimization on the $SE(3)$ manifold resulting in a consistent, unified ecosystem map, depicted as the Finalized Map in Fig.~\ref{fig:Architecture}. The objective (error) function incorporates costs assigned to all landmark to feature point associations and constraints on the keyframe poses based on the localization trajectory (Fig. \ref{fig:Factor_Graph}). The constraints and estimated variables (i.e., keyframe poses and landmark positions) are structured as a locally-connected factor graph to facilitate global optimization~\cite{Dellaert_2021}. The optimization procedure largely follows previous work~\cite{Mur-Artal_2017, Kuemmerle_2011}, the novelty being the constraints imposed by the localization trajectory. 

As depicted in Fig.~\ref{fig:Factor_Graph}, the constraints on the documentation camera keyframe poses derived from the localization trajectory involve two variables, the documentation image acquisition time $t_{i}$ and the transformation $\textbf{T}_{dl}$ between the localization and documentation cameras, both of which are, in turn, constrained by their prior estimates. The documentation and imaging cameras are not required to be precisely synchronized in time, i.e., the acquisition time for each documentation camera image, in terms of the localization trajectory, is only approximately known. Specifically, the constraint on the documentation camera keyframe pose is expressed as a ternary edge in the factor graph wherein $t_{i}$ implies a specific localization camera pose that can be obtained from the trajectory. The localization camera pose can then translated into an implied documentation camera pose via $\textbf{T}_{dl}$ and equation~(\ref{eq:DocPosfromSLAM}). Since the imaging system upon which the cameras are mounted is assumed to be rigid and stable throughout the data collection process, a single value of $\textbf{T}_{dl}$ is estimated for the entire data set. The non-linear optimization attempts to minimize the following error function:
\begin{multline}
    \argmin_{\textbf{X},\textbf{L},\textbf{T}_{dl}, t} \sum\limits_{i,j}\rho_h(r_{i,j}(\textbf{X}_i,L_j)) + \sum\limits_{i}\rho(p_{i}(\textbf{X}_i, t_i,\textbf{T}_{dl} )) \\ + \sum\limits_{i}\rho(t_i-t_{i-obs}) + \rho(\textbf{T}_{dl}^{-1}\textbf{T}_{dl-obs})
\end{multline}
where $\textbf{X}_{i}$ denotes the camera pose, $L_{j}$ the landmark position, $t_{i}$ the documentation image acquisition time, $\textbf{T}_{dl}$ the transformation between the localization and documentation cameras, $\rho$ the squared-error function, $\rho_h$ the robust Huber error function, $r_{i,j}$ the reprojection error for landmark $j$ observed in keyframe $i$ and $p_i$ the pose error between the estimated pose $\textbf{X}_{di}$ of documentation camera $i$ and its pose implied by the localization camera pose $\textbf{X}_{li}$ at the time the documentation image was taken via transformation $\textbf{T}_{dl}$. Specifically:
\begin{equation}
    p_i = \textbf{X}_{di}^{-1}\textbf{X}_{li}\textbf{T}_{dl}
\end{equation}

\vspace{-0.05in}
\section{Experimental Evaluation}

\vspace{-0.05in}
\subsection{Data Collection}

A dual-camera rig was constructed consisting of a Stereolabs ZED-mini interfaced to a Jetson Xavier computer as the localization camera and a Panasonic GH5s configured with a 14 mm prime lens as the documentation camera. The cameras were secured to rigid frame so that their relative orientation was constant. The Panasonic GH5s was oriented downward to provide ecosystem images of the highest quality.  The ZED-mini was either facing directly forward or angled slightly downward ($\approx 25$\textdegree, measured for each deployment). Aligning the localization camera with the direction of travel allows for persistence of features in the FOV and observation of features at a wide range of distances, thereby improving tracking. However, it was found to be advantageous to angle the localization camera slightly downward to observe more proximal features thereby improving motion estimation and avoiding tracking loss. The ZED-mini stereo video was recorded at 60 frames per second (fps) with $1280 \times 720$ resolution and the Panasonic GH5s video at 60 fps with $4096 \times 2160$ resolution.

Data was collected in two visually repetitive environments: a lawn on the University campus and a salt marsh on Sapelo Island, GA, USA. Patches of roughly 10 m $\times$ 10 m were imaged by traversing the area in a lawn-mower (boustrophedon) pattern. At four sites on Sapelo Island, nine AprilTag markers~\cite{olson2011tags} were placed in the imaged patch to serve as ground control points for accuracy assessment. The ground-truth positions of the AprilTags were determined to \textless 2 cm accuracy using an RTK-GPS system (Trimble R12 GNSS receiver with a Trimble TSC7 controller).

\begin{figure}[]
  \includegraphics[width=\linewidth]{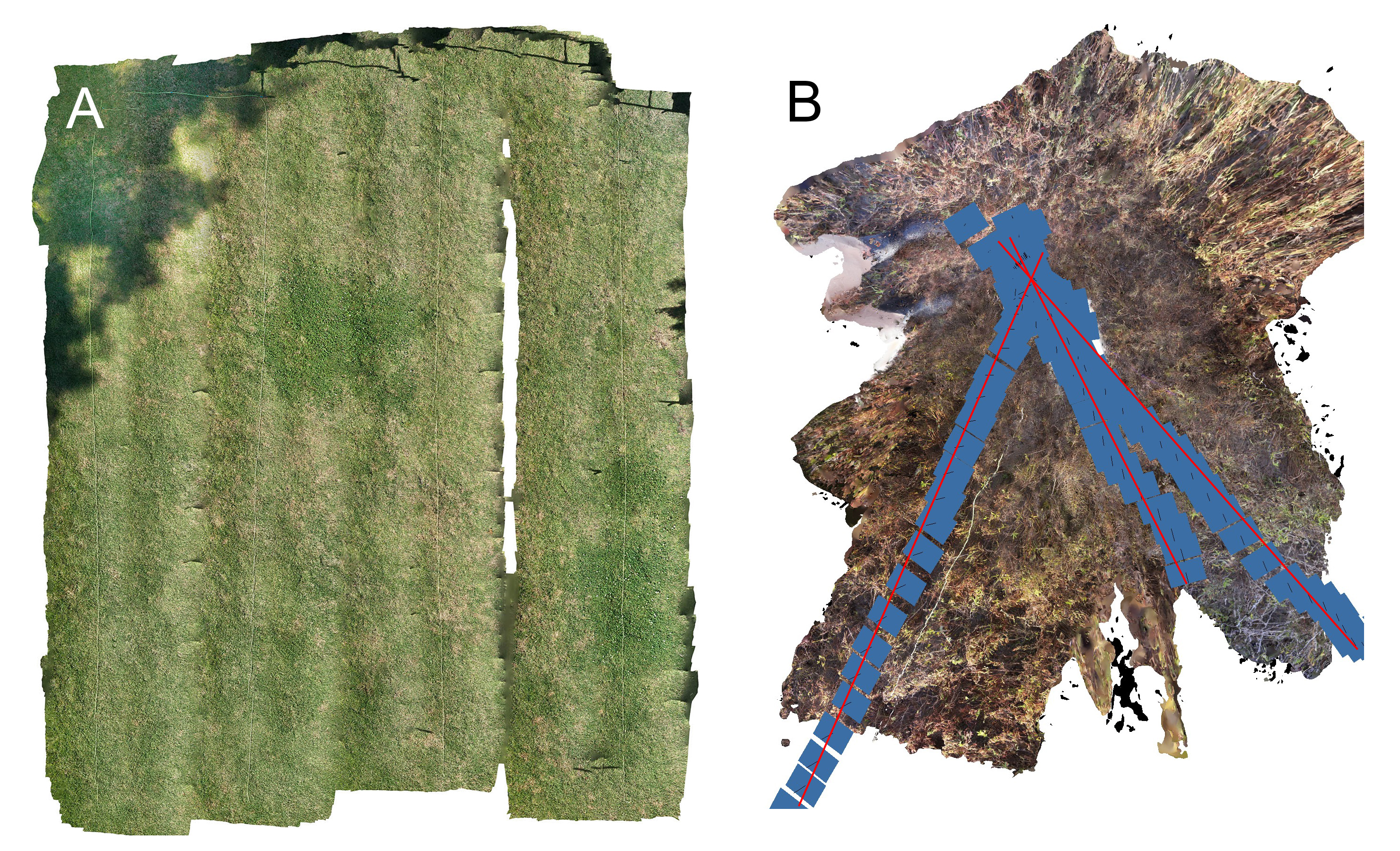}
  \vspace{-0.25in}
  \caption{\textbf{Sample Reconstructions}: A: Accurately reconstructed campus lawn using dual camera SLAM. B: SfM failure due to visual aliasing (blue squares represent aligned images). The three lines of images (highlighted in red) should be parallel but instead converge on a single point in the reconstruction.} 
  \label{fig:sample_recons}
\end{figure}

\vspace{-0.05in}
\subsection{Comparison with SfM}

The dual-camera SLAM approach was compared with two state-of-the-art SfM platforms: COLMAP~\cite{Schonberger_2016} and Agisoft Metashape~\cite{Agisoft_2021}. Six datasets (four from salt marshes on Sapelo Island, two from the campus lawn) were processed to generate maps using the dual-camera SLAM program and the two SfM programs. For this comparison, video frames were extracted at 4 fps from the documentation camera videos (resulting in 80-90\% overlap between frames) and processed using the default SfM program settings that were slightly modified based on preliminary trials to improve reconstruction quality. 
First, the reconstructions were visually assessed to determine if 
the reconstruction was roughly consistent with the planar geometry of the patches and whether the inferred image locations relative to the reconstruction approximately matched the camera trajectory. Second, the completeness of the reconstructions was assessed using the fraction of aligned images as a metric for completeness of the SfM programs and the fraction of visible mesh elements (out of those determined to be potentially visible) as a completeness metric for the dual-camera SLAM system. For the SfM approaches all sub-maps were considered, offering a charitable representation of their performance. On these six datasets, the dual-camera SLAM approach was able to successfully reconstruct repetitive environments in cases when traditional SfM systems either failed entirely or were unable to fully reconstruct the imaged location (Table \ref{table: SfM_Comparison}). COLMAP was better able to generate reconstructions than Metashape, but the reconstructions were typically broken into multiple (up to 22) fragmentary maps. In contrast the dual-camera SLAM approach is able to produce a single, unified map. As examples we show texture mapped reconstructions of a salt marsh grassland (Fig. \ref{fig:Architecture}) and a campus lawn (Fig. \ref{fig:sample_recons}A). The SfM approaches often incorrectly merged subsections due to visual aliasing (Fig. \ref{fig:sample_recons}B).

\begin{table}[ht]
  \begin{center}
    \begin{threeparttable}
      \caption{Comparison with SfM programs}
      \begin{tabular}{c|c|c|c}
          \hline
          Sample  & Metashape & COLMAP & DC-SLAM    \\
          \hline
          P1  &  6\% / Y  & 39\% / Y & 99\% / Y \\
          P2  & 16\% / N  & 92\% / N & 100\% / Y \\
          P3  & 15\% / N  & 65\% / Y & 99\% / Y\\
          P4  & 34\% / Y  & 79\% / Y & 99\% / Y\\
          P5  & 83\% / Y  & 99\% / Y & 99\% / Y \\
          P6  & 40\% / Y  & 100\% / Y & 100\% / Y\\
          \hline
      \end{tabular}
     \begin{tablenotes}
        \small
        \item For each method the percent of images aligned is reported first followed by whether the reconstruction geometry was correct (Y) or not (N).
      \end{tablenotes}
    \label{table: SfM_Comparison}
    \end{threeparttable}
  \end{center}
\end{table}

\begin{figure}[h]
  \makebox[\linewidth][c]{\includegraphics[width=60mm,scale=1.0]{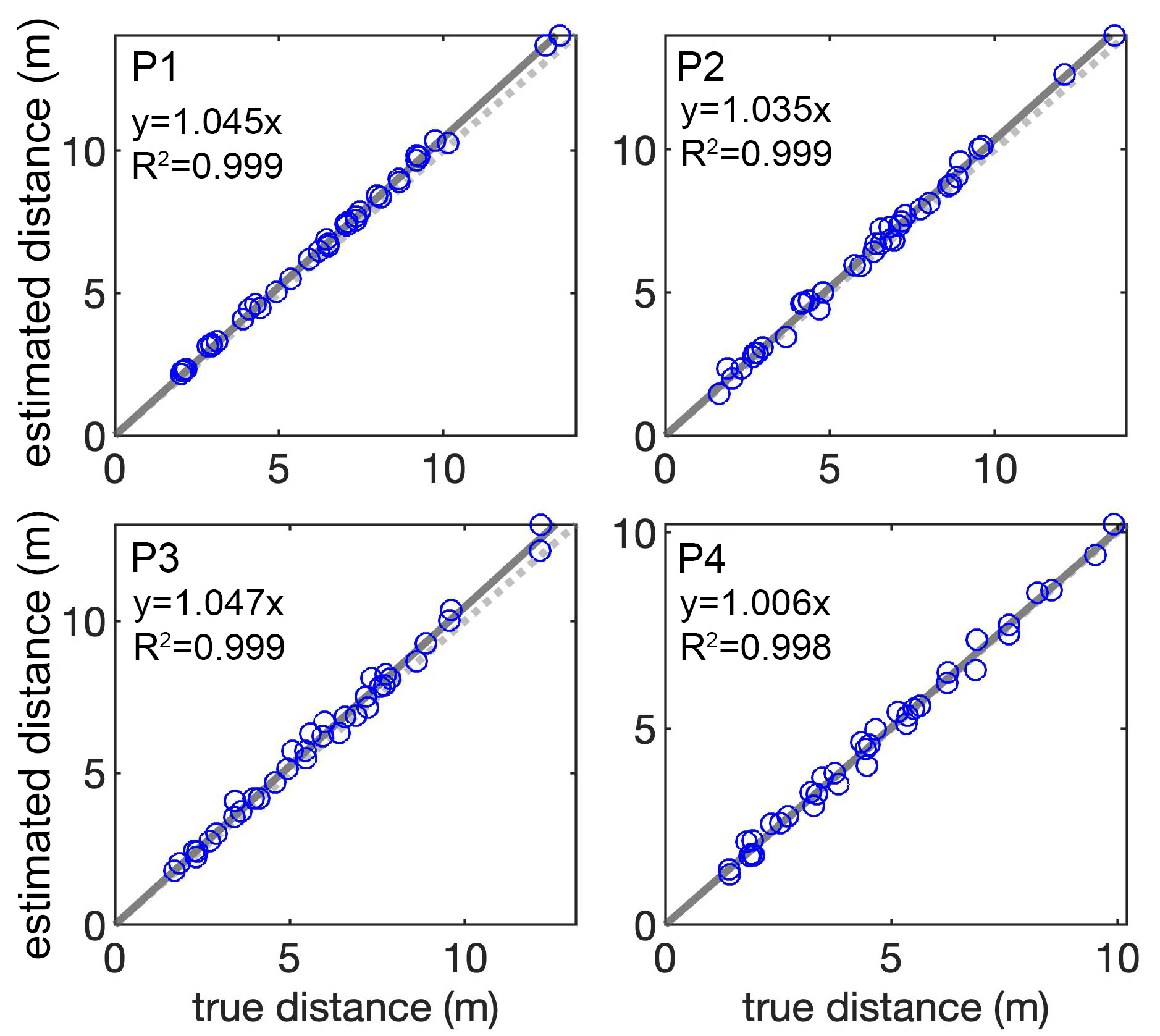}}
  \vspace{-0.2in}
  \caption{Accuracy assessment from inter-tag distances measured using RTK-GPS (true distances) and in dual camera SLAM reconstructions (estimated distances) for four imaged patches (P1-P4). The solid line is a linear regression fit to the data forced through the origin (equation and R\textsuperscript{2} listed on plot)  and the dotted line is the 1:1 line. } \label{fig:Accuracy}
\end{figure}

\vspace{-0.05in}
\subsection{Accuracy of Dual Camera SLAM}

In all the test sites, the dual-camera SLAM system produced reconstructions that appeared reasonable and covered the entire imaged patch (Table \ref{table: SfM_Comparison}). For a more quantitative assessment of the reconstruction quality, distances between ground control points (AprilTags) in the reconstructions were compared with the true distances determined using RTK-GPS positions. At four locations on Sapelo Island, nine AprilTags were placed spanning the imaged patch: four at the corners forming a square defining the edges of the patch, four forming a nested square, and one at the center of the patch.  Determination of AprilTag locations in the reconstructions was done as a post-processing step. After running the dual-camera SLAM program, the reconstructed landmark map from the documentation camera was exported as a point cloud and the documentation camera images and their associated poses were saved. A triangular surface mesh was fit to the point cloud. AprilTags were detected in the documentation camera images and their 3D locations determined via backprojection onto the mesh using the camera poses and inverse camera model. Since most AprilTags were viewed in multiple images, the backprojected 3D positions of all views were averaged to produce a single location for each tag. Euclidean distances between all reconstructed tag pairs and RTK-GPS positions were computed and compared (Fig.~\ref{fig:Accuracy}). The estimated inter-tag distances were generally in good agreement with the true distances (Fig.~\ref{fig:Accuracy}), though slightly overestimated (average 3\%), most likely due to localization camera calibration errors. Nonetheless, the reconstructions were deemed sufficiently accurate for most ecological applications. 

\vspace{-0.05in}
\section{Conclusions and Future Work}
We proposed a dual-camera SLAM approach to map repetitive environments for use in environmental monitoring applications. While the proposed approach does entail a more complicated image acquisition setup, it offers much more reliable mapping of repetitive environments, typical of many ecosystems. Furthermore, decoupling the localization and documentation cameras allows use of cameras ideally suited to each task and flexible swapping of either camera as required for the task at hand. Future improvements include incorporation of additional constraints such as IMU or GPS data~\cite{Forster_2017} into the factor graph for more accurate mapping and development of fully coupled optimization strategies for the localization and documentation camera reconstructions~\cite{Kuo_2020}. 

\section*{Acknowledgements}
\vspace{-0.05in}
This work was supported by the US National Science Foundation (DBI 2016741). We thank the University of Georgia Marine Institute on Sapelo Island for facilitating field work.

\bibliographystyle{IEEEtran}
\bibliography{dualcam_bib}
\end{document}